\PassOptionsToPackage{dvipsnames}{xcolor}

\documentclass{article}

\usepackage{microtype}
\usepackage{graphicx}
\usepackage{subfigure}
\usepackage{booktabs} 

\usepackage{hyperref}

\usepackage[accepted]{icml2025}

\usepackage{amsmath}
\usepackage{amssymb}
\usepackage{mathtools}
\usepackage{amsthm}

\usepackage[capitalize,noabbrev]{cleveref}

\usepackage{lipsum}
\usepackage{pifont}
\usepackage{listings}
\usepackage{subcaption}

\theoremstyle{plain}

\theoremstyle{definition}

\theoremstyle{remark}

\usepackage[textsize=tiny]{todonotes}

\def\eg{\emph{e.g. }}

\def\vs{\emph{vs. }}

\icmltitlerunning{Diffusion Models without Classifier-Free Guidance}

\begin{document}

\twocolumn[
\icmltitle{Diffusion Models without Classifier-free Guidance}



\icmlsetsymbol{equal}{*}

\begin{icmlauthorlist}
\icmlauthor{Zhicong Tang}{thu}
\icmlauthor{Jianmin Bao}{msra}
\icmlauthor{Dong Chen}{msra}
\icmlauthor{Baining Guo}{msra}
\end{icmlauthorlist}


\icmlaffiliation{thu}{Tsinghua University}
\icmlaffiliation{msra}{Microsoft Research Asia}


\icmlkeywords{Diffusion Models, Classifier-free Guidance}

\vskip 0.3in
]



\printAffiliationsAndNotice{}  


\begin{abstract}

This paper presents Model-guidance (MG), a novel objective for training diffusion model that addresses and removes of the commonly used Classifier-free guidance (CFG). Our innovative approach transcends the standard modeling of solely data distribution to incorporating the posterior probability of conditions. The proposed technique originates from the idea of CFG and is easy yet effective, making it a plug-and-play module for existing models. Our method significantly accelerates the training process, doubles the inference speed, and achieve exceptional quality that parallel and even surpass concurrent diffusion models with CFG. Extensive experiments demonstrate the effectiveness, efficiency, scalability on different models and datasets. Finally, we establish state-of-the-art performance on ImageNet $256$ benchmarks with an FID of $1.34$. Our code is available at \href{https://github.com/tzco/Diffusion-wo-CFG}{\texttt{github.com/tzco/Diffusion-wo-CFG}}.

\end{abstract}


\section{Introduction}

Diffusion models~\cite{sohl2015deep,song2019generative,ho2020denoising,songdenoising,songscore} have become the cornerstone of many successful generative models, \eg image generation~\cite{dhariwal2021diffusion,nichol2022glide,rombach2022high,podellsdxl,chenpixart} and video generation~\cite{ho2022video,blattmann2023align,gupta2025photorealistic,polyak2024movie,wang2024lavie} tasks. However, diffusion models also struggle to generate “low temperature" samples~\cite{ho2021classifier,karras2024guiding} due to the nature of training objectives, and techniques such as Classifier guidance~\cite{dhariwal2021diffusion} and Classifier-free guidance (CFG)~\cite{ho2021classifier} are proposed to improve performances.

Despite its advantage and ubiquity, CFG has several drawbacks~\cite{karras2024guiding} and poses challenges to effective implementations~\cite{kynkaanniemi2024applying} of diffusion models. One critical limitation is the simultaneous training of unconditional model apart from the main diffusion model. The unconditional model is typically implemented by randomly dropping the condition of training pairs and replacing with an manually defined empty label. The introduction of additional tasks may reduce network capabilities and lead to skewed sampling distributions~\cite{karras2024guiding,kynkaanniemi2024applying}. Furthermore, CFG requires two forward passes per denoising step during inference, one for the conditioned and another for the unconditioned model, thereby significantly escalating the computational costs.

\begin{figure}[t]
\begin{center}
\centerline{\includegraphics[width=\columnwidth]{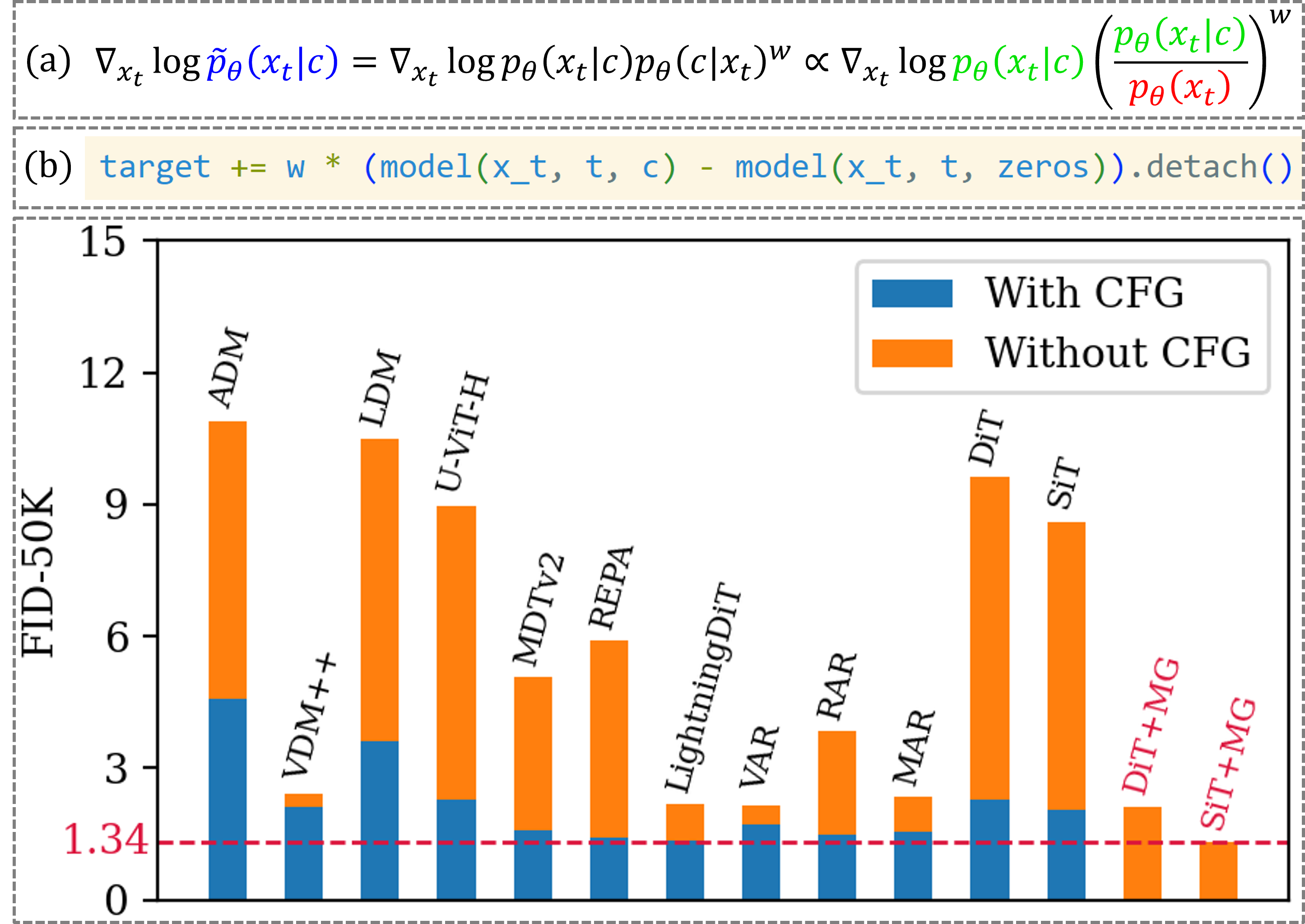}}
\caption{\textbf{We propose Model-guidance (MG), removing Classifier-free guidance (CFG) for diffusion models and achieving state-of-the-art on ImageNet with FID of $\mathbf{1.34}$.} \\
(a) Instead of running models twice during inference (green and red), MG directly learns the final distribution (blue). \\
(b) MG requires only one line of code modification while providing excellent improvements. (c) Comparing to concurrent methods, MG yields lowest FID even without CFG.}
\label{fig:teasor}
\end{center}
\vskip -0.3in
\end{figure}

In this work, we propose Model-guidance (MG), an innovative method for diffusion models to effectively circumvent CFG and boost performances, thereby eliminating the limitations above. We propose a novel objective that transcends from simply modeling the data distribution to incorporating the posterior probability of conditions. Specifically, we leverage the model itself as an implicit classifier and directly learn the score of calibrated distribution during training.

As depicted in~\cref{fig:teasor}, our proposed method confers multiple substantial breakthroughs. First, it significantly refines generation quality and accelerates training processes, with experiments showcasing a $\ge6.5\times$ convergence speedup than vanilla diffusion models with excellent quality. Second, the inference speed is doubled with our method, as each denoising step needs only one network forward in contrast to two in CFG. Besides, it is easy to implement and requires only one line of code modification, making it a plug-and-play module of existing diffusion models with instant improvements. Finally, it is an end-to-end method that excels traditional two-stage distillation-based approaches and even outperforms CFG in generation performances.

We conduct comprehensive experiments on the prevalent Imagenet~\cite{deng2009imagenet,russakovsky2015imagenet} benchmarks with $256\times256$ and $512\times512$ resolution and compare with a wide variates of concurrent models to attest the effectiveness of our proposed method. The evaluation results demonstrate that our method not only parallels and even outperforms other approaches with CFG, but also scales to different models and datasets, making it a promising enhancement for diffusion models. In conclusion, we make the following contribution in this work:

\vspace{-2mm}
\begin{itemize}
    \setlength{\parskip}{0pt}
    \setlength{\itemsep}{2pt}
    \item We proposed a novel and effective method, Model-guidance (MG), for training diffusion models.
    \item MG removes CFG for diffusion models and greatly accelerates both training and inference process.
    \item Extensive experiments with SOTA results on ImageNet demonstrate the usefulness and advantages of MG.
\end{itemize}
\vspace{-2mm}

\begin{figure*}[t]
\vskip -0.2in
\begin{center}
\begin{minipage}[!t]{0.6\linewidth}
\centering
\centerline{\includegraphics[width=1.0\linewidth]{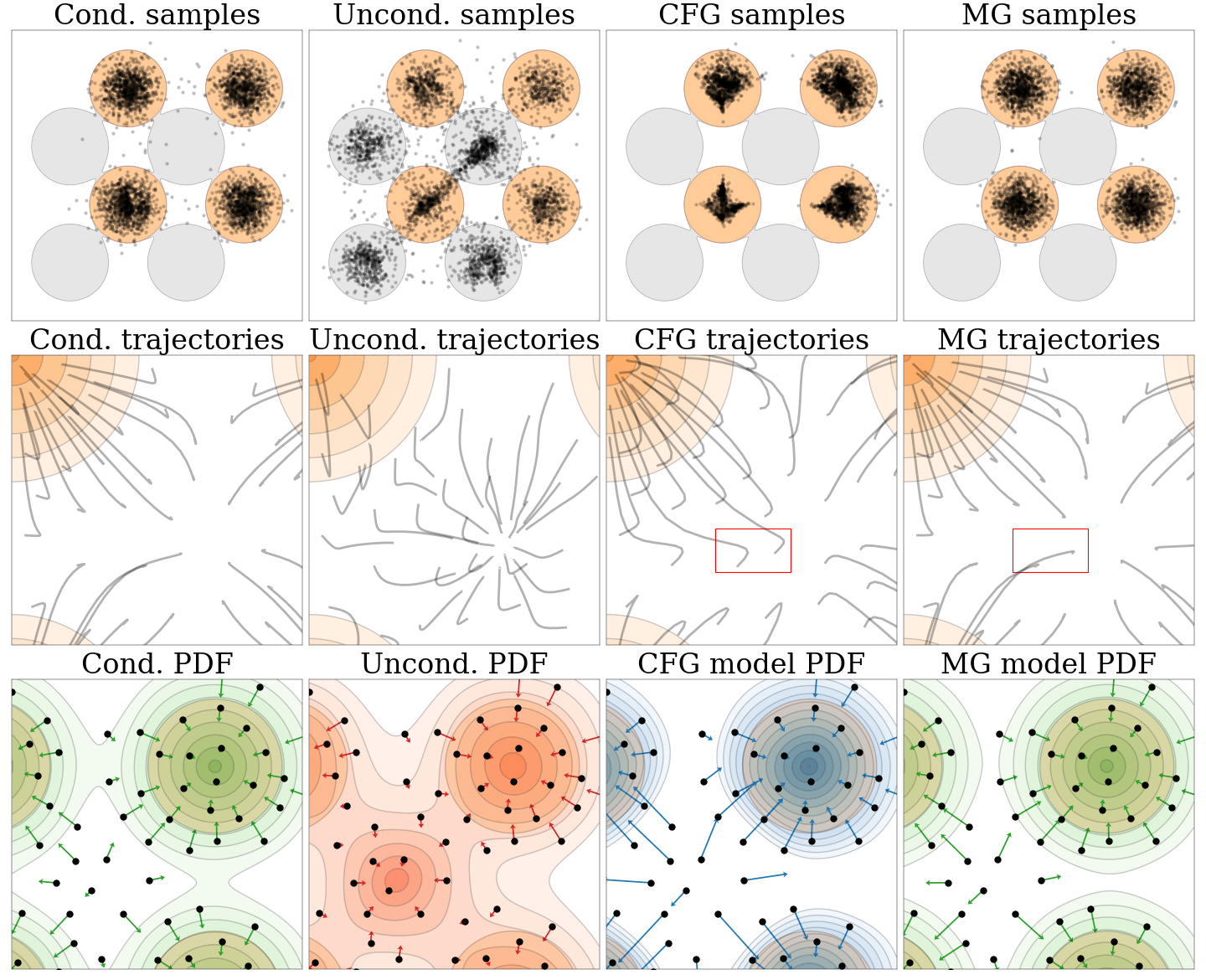}}
\end{minipage}%
\begin{minipage}[!t]{0.4\linewidth}
\vskip 8mm
\caption{We use a grid 2D distribution with two classes, marked with orange and gray regions, as example and train diffusion models on it. We plot the generated samples, trajectories, and probability density function (PDF) of conditional, unconditional, CFG-guided model, and our approach. \\
\textbf{(a)} The first row indicates that although CFG improves quality by eliminating outliers, the samples concentrate in the center of data distributions, resulting the loss of diversity. In contrast, our method yields less outliers than the conditional model and a better coverage of data than CFG. \\
\textbf{(b)} In the second row, the trajectories of CFG show sharp turns at the beginning, \eg samples inside the red box, while our method directly drives the samples to the closet data distributions. \\
\textbf{(c)} The PDF plots of the last row also suggest that our method predicts more symmetric contours than CFG, balancing both quality and diversity. \\
}
\end{minipage}
\label{fig:toy-example}
\end{center}
\vskip -0.3in
\end{figure*}


\abovedisplayskip=6pt
\belowdisplayskip=6pt

\section{Background}

\subsection{Diffusion and Flow Models}

\textbf{Diffusion models}~\cite{sohl2015deep,song2019generative,ho2020denoising,songdenoising,songscore} are a class of generative models that utilize forward and reverse stochastic processes to model complex data distributions.

The forward process adds noise and transforms data samples into Gaussian distributions as
\begin{equation}
\label{eq:diffusion_marginal}
q(x_t | x_0) = \mathcal{N}\left(x_t; \sqrt{\bar{\alpha}_t} x_0, (1 - \bar{\alpha}_t)\mathbf{I}\right),
\end{equation}
where $x_t$ represents the noised data at timestep $t$ and $\bar{\alpha}_t = \prod_{s=1}^t \alpha_s$ is the noise schedule.

Conversely, the reverse process learns to denoise and finally recover the original data distribution, which aims to reconstruct score~\cite{sohl2015deep,songscore} from the noisy samples $x_t$ by learning
\begin{equation}
p_\theta(x_{t-1} | x_t) = \mathcal{N}\left(x_{t-1}; \mu_\theta(x_t, t), \Sigma_\theta(x_t, t)\right),
\end{equation}
where $\mu_\theta$ and $\Sigma_\theta$ are mean and variance and commonly predicted by neural networks.

In common implementations, the training of diffusion models leverages a re-parameterized objective that directly predicts the noise at each step~\cite{ho2020denoising}
\begin{equation}
\mathcal{L}_{\text{simple}} = \mathbb{E}_{t, x_0, \epsilon} \| \epsilon_\theta(x_t, t) - \epsilon \|^2 ,
\label{eq:diffusion_loss_simple}
\end{equation}
where $x_t$ is derived from the forward process in~\cref{eq:diffusion_marginal} with $x_0$ and $\epsilon$ drawn from dataset and Gaussian noises.

Conditional diffusion models allow users to generate samples aligned with specified demands and precisely control the contents of samples. In this case, the generation process is manipulated with give conditions $c$, such as class labels or text prompts, where network functions are $\epsilon_\theta(x_t,t,c)$.

\textbf{Flow Models}~\cite{lipmanflow,liuflow,albergo2023stochastic,tongimproving} are another emerging type of generative models similar to diffusion models. Flow models utilize the concept of Ordinary Differential Equations (ODEs) to bridge the source and target distribution and learn the directions from noise pointing to ground-truth data.

The forward process of flow models is defined as an Optimal Transport (OT) interpolant~\cite{mccann1997convexity}
\begin{equation}
    x_t = (1-t)x_0+t\epsilon,
\end{equation}
and the loss function takes the form~\cite{lipmanflow}
\begin{equation}
    \mathcal{L}_\textup{FM} = \mathbb{E}_{t,x_0,\epsilon}\left\|u_\theta(x_t)-u_t(x_t|x_0)\right\|^2,
\end{equation}
where the ground-truth conditional flow is given by
\begin{equation}
    \label{eq:ground-truth-flow}
    u_t(x_t|x_0)=x_0-\epsilon.
\end{equation}

\subsection{Classifier-Free Guidance}

Classifier-free guidance (CFG)~\cite{ho2021classifier} is a widely adopted technique in conditional diffusion models to enhance generation performance and alignment to conditions. It provides an explicit control of the focus on conditioning variables and avoids to sample within the “low temperature" regions with low quality.

The key design of CFG is to combine the posterior probability and utilize Bayes' rule during inference time. To facilitate this, it is required to train both conditional and unconditional diffusion models. In particular, CFG trains the models to predict
\begin{align}
&\epsilon_\theta(x_t, t, c) \propto -\nabla_{x_t} \log p_\theta(x_t | c), \\
&\epsilon_\theta(x_t, t, \varnothing) \propto - \nabla_{x_t} \log p_\theta(x_t),
\end{align}
\noindent where is an additional empty class introduced in common practices. During training, the model switches between the two modes with a ratio $\lambda$.

For inference, the model combines the conditional and unconditional scores and guides the denoising process as
\begin{equation}
\label{eq:cfg_mod_noise}
\tilde{\epsilon}_\theta(x_t, t, c) = \epsilon_\theta(x_t, t, c) + w \cdot \left( \epsilon_\theta(x_t, t, c) - \epsilon_\theta(x_t, t, \varnothing) \right),
\end{equation}
\noindent where $w$ is the guidance scale that controls the focus on conditional scores and the trade-off between generation performance and sampling diversity. CFG has become an widely adopted protocol in most of diffusion models for tasks, such as image generation and video generation.

\subsection{Distillation-based Methods}

Besides acceleration~\cite{song2023consistency}, researchers~\cite{sauer2024fast} also adopt distillation on diffusion models with CFG to improve sampling quality. Rectified Flow~\cite{liuflow} disentangles generation trajectories and streamline learning difficulty by alternatively using offline model to provide training pairs for online models. 
Distillation is also used to learn a smaller one-step model to match the generation performance of larger multi-step models~\cite{meng2023distillation}. Pioneering diffusion models~\cite{flux,stablediffusion35} are released with a distillated version, where CFG scale is viewed as an additional embedding to provide accurate control. However, these approaches involve two-stage learning and require extra computation and storage for offline teacher models.


\section{Method}

\subsection{Rethinking Classifier-free guidance}

Due to the complex nature of visual datasets, diffusion models often struggle whether to recover real image distribution or engage in the alignment to conditions. Classifier-free guidance (CFG) is then proposed and has become an indispensable ingredient of modern diffusion models~\cite{nichol2021improved,karras2022elucidating,saharia2022photorealistic,hoogeboom2023simple}. It drives the sample towards the regions with higher likelihood of conditions with \cref{eq:cfg_mod_noise}, where the images are more canonical and better modeled by networks~\cite{karras2024guiding}.

However, CFG has with several disadvantages~\cite{karras2024guiding,kynkaanniemi2024applying}, such as the multitask learning of both conditional and unconditional generation, and the doubled number of function evaluations (NFEs) during inference. Moreover, the tempting property that solving the denoising process according to \cref{eq:cfg_mod_noise} eventually recovers data distribution does not hold, as the joint distribution does not represent a valid heat diffusion of the ground-truth~\cite{zheng2024characteristicguidancenonlinearcorrection}. This results in exaggerated truncation and mode dropping similar to~\cite{Karras2018ASG,brock2019largescalegantraining,Sauer2022StyleGANXLSS}, since the samples are blindly pushed towards the regions with higher posterior probability. The generation trajectories are distorted in~\cref{fig:toy-example}, the images are often over-saturated in color, and the content of samples is overly simplified.

CFG originates from the classifier-guidance~\cite{dhariwal2021diffusion} that incorporates an auxiliary classifier model $p_\theta(c|x_t)$ to modify the sampling distribution as
\begin{equation}
\tilde{p}_\theta(x_t|c)\propto p_\theta(x_t|c)p_\theta(c|x_t)^w,
\end{equation}
\noindent and estimates the posterior probability term with Bayes' rule
\begin{equation}
p_\theta(c|x_t)=\frac{p_\theta(x_t|c)p_\theta(c)}{p_\theta(x_t)},
\label{eq:cfg_bayes}
\end{equation}
\noindent where $p_\theta(x_t|c)$ and $p_\theta(x_t)$ are conditional and unconditional distributions, respectively.

The unconditional model is usually implemented by randomly replacing labels by an empty class with a ratio $\lambda$. During inference, each sample is typically forwarded twice, one with and one without conditions. The finding naturally leads us to the question: can we fuse the auxiliary classifier into diffusion models in a more \textit{efficient} and \textit{elegant} way?

\setcounter{figure}{0}
\begin{figure}[t]
\vskip -2mm
    \centering
    \begin{minipage}[!t]{0.5\linewidth}
        \centering
        \includegraphics[width=1\textwidth]{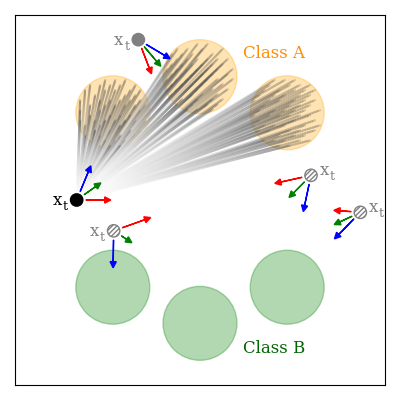}
        \vskip -5mm
        \captionsetup{labelformat=empty,width=0.92\textwidth}
        \caption{\fontsize{8pt}{8pt}\selectfont (a) \textcolor{red}{Unconditional}, \textcolor{ForestGreen}{Conditional}, and \textcolor{blue}{Classifier-free Guided} score.}
        \vspace{1mm}
    \end{minipage}%
    \begin{minipage}[!t]{0.5\linewidth}
        \centering
        \includegraphics[width=0.96\textwidth]{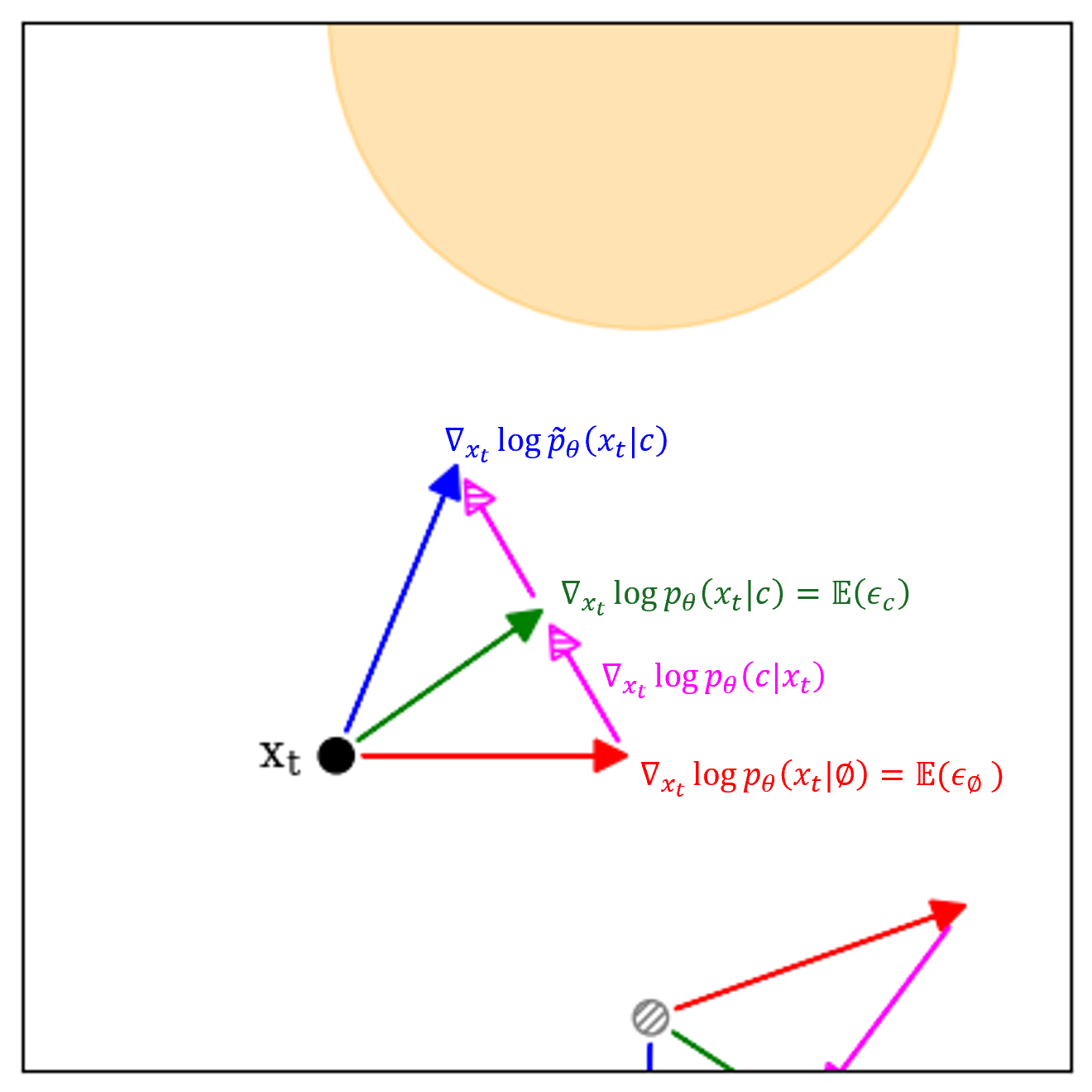}
        \vskip -4mm
        \captionsetup{labelformat=empty,width=0.8\textwidth}
        \caption{\fontsize{8pt}{8pt}\selectfont (b) \textcolor{magenta}{The offsets of CFG} push update directions to the data.}
    \end{minipage}
    \vskip -3mm
\caption{Illustration of our method. (a) The green and red arrow point towards the centroids of data distributions, as the training pairs $(x_0, \epsilon)$ are randomly sampled. (b) While CFG provides accurate directions by subtracting the two vectors, our method directly learns the blue arrow, $\nabla\log \tilde{p}_\theta(x_t|c)$.}
\label{fig:method}
\vskip -0.15in
\end{figure}

\subsection{Model-guidance Loss}

Conditional diffusion models optimize the conditional probability $p_\theta(x_t|c)$ by \cref{eq:diffusion_loss_simple}, where $x_t$ is the noisy data and $c$ is the condition, \eg, labels and prompts. However, the models tend to ignore the condition in common practices and CFG~\cite{ho2020denoising} is proposed as an explicit bias.

To enhance both generation quality and alignment to conditions, we propose to take into account the posterior probability $p_\theta(c|x_t)$. This leads to the joint optimization of $\tilde{p}_\theta(x_t|c)=p_\theta(x_t|c)p_\theta(c|x_t)^w$, where $w$ is the weighting factor of posterior probability. The score of the joint distribution is formulated as
\begin{equation}
\nabla_{x_t}\log\tilde{p}_\theta(x_t|c)=\nabla_{x_t}\log p_\theta(x_t|c)+w\cdot\nabla_{x_t}\log p_\theta(c|x_t)
\label{eq:joint_score}
\end{equation}
The first term corresponds to the standard diffusion objective in \cref{eq:diffusion_loss_simple}. However, the second term represents the score of posterior probability $p_\theta(c|x_t)$ and cannot be directly obtained, since an explicit classifier of noisy samples is unavailable. Inspired by \cref{eq:cfg_bayes}, we transform the diffusion model into an implicit classifier and let it guide itself. Specifically, we employ Bayes' rule to estimate
\begin{align}
\label{eq:bayes_decomp}
\log p_\theta(c|x_t) &= \log p_\theta(x_t|c) - \log p_\theta(x_t) + \log p_\theta(c) \notag \\
            &\propto \log p_\theta(x_t|c) - \log p_\theta(x_t) \quad
\end{align}
Next, we use the diffusion model to approximate the scores
\begin{align}
\label{eq:cond_score}
& \nabla_{x_t} \log p_t(x_t|c) = -\frac{1}{\sigma_t}\epsilon_\theta(x_t,t,c), \\
\label{eq:uncond_score}
& \nabla_{x_t} \log p_t(x_t) = -\frac{1}{\sigma_t}\epsilon_\theta(x_t,t,\varnothing),
\end{align}
where $\sigma_t$ is the variance of the noise added to $x_t$ at timestep $t$, $\varnothing$ is the empty class, and $\epsilon_\theta(\cdot)$ is the diffusion model. Substituting \cref{eq:cond_score,eq:uncond_score} into \cref{eq:bayes_decomp} yields the score of posterior probability
\begin{equation}
   \nabla_{x_t} \log p_\theta(c|x_t) \propto \frac{1}{\sigma_t}\left(\epsilon_\theta(x_t,t,\varnothing) - \epsilon_\theta(x_t,t,c)\right).
    \label{eq:score_diff}
\end{equation}
Then, our method applies the Bayes' estimation in \cref{eq:bayes_decomp} online and trains a conditional diffusion model to directly predict the score in \cref{eq:joint_score}, instead of separately learning \cref{eq:cond_score,eq:uncond_score} in the form of CFG. A straight-forward implementation is to adopt the objective in \cref{eq:diffusion_loss_simple} with a modified optimization target
\begin{align}
\label{eq:final_loss}
&\mathcal{L}_{\text{MG}} = \mathbb{E}_{t,(x_0,c),\epsilon}\| \epsilon_\theta(x_t,t,c) - \epsilon' \|^2, \\
\label{eq:mg_noise}
&\epsilon' = \epsilon + w\cdot\text{sg}(\tilde{\epsilon}_\theta(x_t,t,c) - \tilde{\epsilon}_\theta(x_t,t,\varnothing)).
\end{align}
\noindent We apply the stop gradient operation, $\text{sg}(\cdot)$, which is a common practice of avoiding model collapse~\cite{grill2020bootstrap}. We also use the Exponential Mean Average (EMA) counterpart of the online model, $\tilde{\epsilon}_\theta(\cdot)$, to stabilize the training process and provide accurate estimations. For flow-based models, we have the similar objective
\begin{align}
\label{eq:final_flow_loss}
&\mathcal{L}_{\text{MG}} = \mathbb{E}_{t,(x_0,c),\epsilon}\| u_\theta(x_t,t,c) - u' \|^2, \\
\label{eq:mg_flow}
&u' = u + w\cdot\text{sg}(u_\theta(x_t,t,c) - u_\theta(x_t,t,\varnothing)).
\end{align}
\noindent where $u$ is the ground-truth flow in \cref{eq:ground-truth-flow}.

\begin{algorithm}[t]
   \caption{Training with Model-guidance Loss}
   \label{alg:training-mg}
    \begin{algorithmic}
       \STATE {\bfseries Input:} dataset $\{\mathbf{X_i},\mathbf{C_i}\}$, noise schedule $\bar{\alpha}$, model $\epsilon_\theta$
       \REPEAT
       \STATE Sample data $(x_0, c)\sim\{\mathbf{X_i},\mathbf{C_i}\}$
       \STATE Sample noise $\epsilon\sim\mathcal{N}(0,1)$ and time $t\sim\mathbf{U}(0,1)$
       \STATE Add noise with $x_t= \sqrt{\bar{\alpha}_t}x_0+ \sqrt{1-\bar{\alpha}_t}\epsilon$
       \STATE \textcolor{blue}{Modify target $\epsilon' = \epsilon + w\cdot\text{sg}(\epsilon_\theta(x_t,c,t) - \epsilon_\theta(x_t,\varnothing,t))$}
       \STATE Compute loss $\mathcal{L}_{\text{MG}} = \| \epsilon_\theta(x_t,c,t) - \epsilon' \|^2$
       \STATE Back propagation $\theta = \theta - \eta \nabla_\theta \mathcal{L}_{\text{MG}}$
       \UNTIL{converged}
    \end{algorithmic}
\end{algorithm}

During training, we randomly drop the condition $c$ in \cref{eq:final_loss,eq:final_flow_loss} to $\varnothing$ with a ratio of $\lambda$. These formulations transform the model itself into an implicit classifier and adjust the standard training objective of diffusion model in a self-supervised manner, allowing the joint optimization of generation quality and condition alignment with the minimum modification of existing pipelines.

\subsection{Implementation Details}

With the MG formulation in \cref{eq:final_loss,eq:final_flow_loss}, we have adequate options in the detailed implementations, such as incorporating an additional input of the guidance scale $w$ into networks, replacing the usage of empty class with the law of total probability, and whether to manual or automatically adjust the hyper-parameters.

\textbf{Scale-aware networks.} Similar to other distillation-based methods~\cite{frans2024one}, the guidance scale $w$ can be fed into the network as an additional condition. When augmented with $w$-input, our models offer flexible choices of the balance between image quality and sample diversity during inference time. Note that our models require only one forward per step for all values of $w$, while standard CFG needs two forwards, \eg, one with condition and one without condition. In particular, we sample guidance scale from an specified interval, and the loss function are modified into the following form
\begin{align}
    &\mathcal{L}_{\text{MG}} = \mathbb{E}_{t,(x_0,c),\epsilon,w}\| \epsilon_\theta(x_t,t,c,w) - \epsilon' \|^2, \\
    &\epsilon' = \epsilon + w\cdot\text{sg}(\epsilon_\theta(x_t,t,c,1) - \epsilon_\theta(x_t,t,\varnothing,0)).
    \label{eq:scale_input_loss}
\end{align}
\textbf{Removing the empty class.} Another option is whether to perform multitask learning of both conditional and unconditional generation with the same model. In CFG, the estimator in \cref{eq:cfg_bayes} requires to train an unconditional model. However, the multitask learning can distract and hinder model capability. Using the law of total probability
\begin{align}
    \nabla_{x_t} \log p_t(x_t) &= \nabla_{x_t} \log \sum_{c} p_t(x_t|c)p_t(c) \notag \\
    &= -\frac{1}{N\sigma_t}\sum_{i=1}^{N}\epsilon_\theta(x_t,t,c_i),
\end{align}
where $N$ different labels are used to estimate the unconditional score, our models focus on the conditional prediction and avoid the introduction of additional empty class.

\textbf{Automatic adjustment of the hyper-parameter $w$.} While the scale $w$ in \cref{eq:mg_noise,eq:mg_flow} plays an important role, it is tedious and costly to perform manual search during training. Therefore, we introduce an automatic scheme to adjust $w$. We begin with $w=0$ that corresponds to vanilla diffusion models, then update the value with EMA according to intermediate evaluation results. The value of $w$ is raised when quality decreases and suppressed otherwise, leading to an optimums when the training converged.


\section{Experiment}

We first present a system-level comparison with state-of-the-art models on ImageNet $256\times256$ conditional generation. Then we conduct ablation experiments to investigate the detained designs of our method. Especially, we emphasize on the following questions:

\vspace{-4mm}
\begin{itemize}
    \setlength{\parskip}{0pt}
    \setlength{\itemsep}{2pt}
    \item How far can MG push the performances of existing diffusion models? (\cref{tab:imagenet-256-wo-cfg,tab:imagenet-256-w-cfg}, \cref{sec:overall_performance})
    \item How does implementation details influence the gain of proposed method? (\cref{tab:ablation-scale,tab:ablation-drop-ratio,tab:ablation-input-scale,tab:ablation-empty-class}, \cref{sec:ablation_study})
    \item Can MG scales to larger models and datasets with efficiency? (\cref{tab:model-size,tab:imagenet-512}, \cref{fig:fid-vs-cfg-scale,fig:training-speedup,fig:parameters-inference-speedup}, \cref{sec:ablation_study})
\end{itemize}
\vspace{-4mm}

\begin{table}[t]
\caption{Experiments on ImageNet 256 without CFG. By deploying our method, the performances of both DiT-XL/2 and SiT-XL/2 are greatly boosted, achieving state-of-the-art.}
\label{tab:imagenet-256-wo-cfg}
\begin{center}
\begin{small}
\begin{sc}
\vskip -0.2in
\scalebox{0.8}{
\begin{tabular}{l|ccccc|c}
\toprule
Model & FID$\downarrow$   & sFID$\downarrow$  & IS$\uparrow$     & Pre.$\uparrow$ & Rec.$\uparrow$ & img/s$\uparrow$ \\
\midrule
ADM & 10.9 & - & 101.0 & 0.69 & 0.63 & - \\
VDM++ & 2.40 & - & 225.3 & 0.78 & 0.66 & - \\
\midrule
LDM-4 & 10.5 & - & 103.5 & 0.71 & 0.62 & - \\
U-ViT-H & 8.97 & - & 136.7 & 0.69 & 0.63 & - \\
MDTv2 & 5.06 & - & 155.6 & 0.72 & 0.66 & 0.2 \\
REPA & 5.90 & 6.33 & 162.1 & 0.71 & 0.56 & 0.76 \\
L-DiT & 2.17 & 4.36 & 205.6 & 0.77 & 0.65 & 0.06 \\
\midrule
VAR$_{d30}$ & 2.16 & - & 288.7 & 0.81 & 0.61 & 11.2 \\
RAR$_\textup{XXL}$ & 3.83 & - & 274.5 & 0.79 & 0.61 & 3.9 \\
MAR-H & 2.35 & - & 227.8 & 0.79 & 0.62 & 0.6 \\
\midrule
DiT-XL/2 & 9.62 & 6.85 & 121.5 & 0.67 & 0.67 & 0.2 \\
+MG$_\textup{(ours)}$ & \underline{\textbf{2.03}} & 4.36 & 292.1 & 0.81 & 0.66 & 0.2 \\
\color{gray}{Improve} & \color{ForestGreen}{$78.9\%$} & \color{ForestGreen}{$36.4\%$} & \color{ForestGreen}{$140\%$} & \color{ForestGreen}{$20.9\%$} & \color{OrangeRed}{$1.49\%$} & \color{ForestGreen}{$0.0\%$} \\
\midrule
SiT-XL/2 & 8.61 & 6.32 & 131.7 & 0.68 & 0.67 & 0.76 \\
+MG$_\textup{(ours)}$ & \underline{\textbf{1.34}} & 4.58 & 321.5 & 0.81 & 0.65 & 0.76 \\
\color{gray}{Improve} & \color{ForestGreen}{$84.4\%$} & \color{ForestGreen}{$27.5\%$} & \color{ForestGreen}{$144\%$} & \color{ForestGreen}{$19.1\%$} & \color{OrangeRed}{$2.99\%$} & \color{ForestGreen}{$0.0\%$} \\
\bottomrule
\end{tabular}
}%
\end{sc}
\end{small}
\end{center}
\vskip -0.1in
\end{table}

\begin{table}[t]
\caption{Experiments on ImageNet 256 with CFG. Comparing to models with CFG, our method still obtains excellent results and surpasses others without efficiency loss.}
\label{tab:imagenet-256-w-cfg}
\begin{center}
\begin{small}
\begin{sc}
\vskip -0.2in
\scalebox{0.8}{
\begin{tabular}{l|ccccc|c}
\toprule
Model & FID$\downarrow$   & sFID$\downarrow$  & IS$\uparrow$     & Pre.$\uparrow$ & Rec.$\uparrow$ & img/s$\uparrow$ \\
\midrule
ADM & 4.59 & 5.25 & 186.7 & 0.82 & 0.52 & - \\
VDM++ & 2.12 & - & 267.7 & 0.81 & 0.65 & - \\
\midrule
LDM & 3.60 & - & 247.7 & 0.87 & 0.48 & - \\
U-ViT-H & 2.29 & 5.68 & 263.9 & 0.82 & 0.57 & - \\
MDTv2 & 1.58 & 4.52 & 314.7 & 0.79 & 0.65 & 0.1 \\
REPA & 1.42 & 4.70 & 305.7 & 0.80 & 0.65 & 0.39 \\
L-DiT & 1.35 & 4.15 & 295.3 & 0.79 & 0.65 & 0.03 \\
\midrule
VAR$_{d30}$ & 1.73 & - & 350.2 & 0.82 & 0.60 & 6.3 \\
RAR$_\textup{XXL}$ & 1.48 & - & 326.0 & 0.80 & 0.63 & 2.1 \\
MAR-H & 1.55 & - & 303.7 & 0.81 & 0.62 & 0.3 \\
\midrule
DiT-XL/2 & 2.27 & 4.60 & 278.2 & 0.83 & 0.57 & 0.1 \\
+MG$_\textup{(ours)}$ & \underline{\textbf{2.03}} & 4.36 & 292.1 & 0.81 & 0.66 & 0.2 \\
\color{gray}{Improve} & \color{ForestGreen}{$10.6\%$} & \color{ForestGreen}{$5.22\%$} & \color{ForestGreen}{$5.00\%$} & \color{OrangeRed}{$2.41\%$} & \color{ForestGreen}{$15.8\%$} & \color{ForestGreen}{$100\%$} \\
\midrule
SiT-XL/2 & 2.06 & 4.49 & 277.5 & 0.83 & 0.59 & 0.39 \\
+MG$_\textup{(ours)}$ & \underline{\textbf{1.34}} & 4.58 & 321.5 & 0.81 & 0.65 & 0.76 \\
\color{gray}{Improve} & \color{ForestGreen}{$35.0\%$} & \color{OrangeRed}{$2.00\%$} & \color{ForestGreen}{$15.9\%$} & \color{OrangeRed}{$2.41\%$} & \color{ForestGreen}{$10.2\%$} & \color{ForestGreen}{$94.9\%$} \\
\bottomrule
\end{tabular}
}%
\end{sc}
\end{small}
\end{center}
\vskip -0.1in
\end{table}

\subsection{Setup}

\textbf{Implementation and dataset.} We follow the experiment pipelines in DiT~\cite{peebles2023scalable} and SiT~\cite{ma2024sit}. We use ImageNet~\cite{deng2009imagenet,russakovsky2015imagenet} dataset and the Stable Diffusion~\cite{rombach2022high} VAE to encode $256\times256$ images into the latent space of $\mathbb{R}^{32\times32\times4}$. We conduct ablation experiments with the B/2 variant of DiT and SiT models and train for 400K iterations. During training, we use AdamW~\cite{kingma2014adam,loshchilov2017decoupled} optimizer and a batch size of 256 in consistent with DiT~\cite{peebles2023scalable} and SiT~\cite{ma2024sit} for fair comparisons. For inference, we use 1000 sampling steps for DiT models and Euler-Maruyama sampler with 250 steps for SiT.

\textbf{Baseline Models.} We compare with several state-of-the-art image generation models, including both diffusion-based and AR-based methods, which can be classified into the following three classes: (a) \textit{Pixel-space diffusion}: ADM~\cite{dhariwal2021diffusion}, VDM++~\cite{kingma2023understanding}; (b) \textit{Latent-space diffusion}: LDM~\cite{rombach2022high}, U-ViT~\cite{bao2023all}, MDTv2~\cite{gao2023masked}, REPA~\cite{yu2024representation}, LightningDiT(L-DiT)~\cite{yao2025reconstruction}, DiT~\cite{peebles2023scalable}, SiT~\cite{ma2024sit}; (c) \textit{Auto-regressive models}: VAR~\cite{tian2024visual}, RAR~\cite{yu2024randomized}, MAR~\cite{li2024autoregressive}. These models consist of strong baselines and demonstrate the advantages of our method. Although our method does not requires CFG during inference, we still compare with these baselines under two settings, with and without CFG, for thoroughly investigations. 

\textbf{Evaluation metrics.} We report the commonly used Frechet inception distance~\cite{heusel2017gans} with 50,000 samples (FID-50K). In addition, we report sFID~\cite{nash2021generating}, Inception Score (IS)~\cite{salimans2016improved}, Precision (Pre.), and Recall (Rec.)~\cite{kynkaanniemi2019improved} as supplementary metrics. We also report the time to generate one sample of each model in seconds to measure the trade-off between generation quality and computation budget.

\begin{table}[t]
\caption{Experiments on scale $w$.}
\label{tab:ablation-scale}
\begin{center}
\begin{small}
\begin{sc}
\vskip -0.2in
\scalebox{0.8}{
\begin{tabular}{lc|ccccc}
\toprule
Model & $w$ & FID$\downarrow$   & sFID$\downarrow$  & IS$\uparrow$     & Pre.$\uparrow$ & Rec.$\uparrow$ \\
\midrule
DiT-B/2 & 1.00 & 43.5 & 36.7 & 39.23 & 0.62 & 0.34 \\
+MG$_\textup{(ours)}$ & 1.25 & 9.86 & 8.87 & 176.1 & 0.81 & 0.37 \\
+MG$_\textup{(ours)}$ & 1.50 & \underline{\textbf{7.24}} & \underline{\textbf{5.56}} & 189.2 & 0.84 & 0.38 \\
+MG$_\textup{(ours)}$ & 1.75 & 8.21 & 6.63 & 197.2 & \underline{\textbf{0.86}} & 0.38 \\
+MG$_\textup{(ours)}$ & 2.00 & 9.66 & 7.90 & \underline{\textbf{224.7}} & 0.85 & \underline{\textbf{0.39}} \\
+MG$_\textup{(ours)}$ & Auto & 7.60 & 6.29 & 192.4 & 0.85 & 0.38 \\
\midrule
SiT-B/2 & 1.00 & 33.0 & 27.8 & 65.24 & 0.68 & 0.35 \\
+MG$_\textup{(ours)}$ & 1.25 & 8.94 & 7.87 & 194.3 & 0.83 & 0.38 \\
+MG$_\textup{(ours)}$ & 1.50 & \underline{\textbf{6.49}} & \underline{\textbf{5.69}} & 212.3 & 0.86 & 0.38 \\
+MG$_\textup{(ours)}$ & 1.75 & 8.03 & 6.91 & 221.0 & 0.86 & 0.39 \\
+MG$_\textup{(ours)}$ & 2.00 & 9.14 & 7.99 & \underline{\textbf{236.7}} & \underline{\textbf{0.88}} & \underline{\textbf{0.40}} \\
+MG$_\textup{(ours)}$ & Auto & 6.86 & 5.88 & 219.1 & 0.87 & 0.38 \\
\bottomrule
\end{tabular}
}%
\end{sc}
\end{small}
\end{center}
\vskip -0.1in
\end{table}

\begin{table}[t]
\caption{Experiments on drop ratio $\lambda$.}
\label{tab:ablation-drop-ratio}
\begin{center}
\begin{small}
\begin{sc}
\vskip -0.2in
\scalebox{0.8}{
\begin{tabular}{lc|ccccc}
\toprule
Model & $\lambda$ & FID$\downarrow$   & sFID$\downarrow$  & IS$\uparrow$     & Pre.$\uparrow$ & Rec.$\uparrow$ \\
\midrule
DiT-B/2 & 1.00 & 43.5 & 36.7 & 39.23 & 0.62 & 0.34 \\
+MG$_\textup{(ours)}$ & 0.05 & 11.7 & 9.90 & 156.7 & 0.78 & 0.33 \\
+MG$_\textup{(ours)}$ & 0.10 & \underline{\textbf{7.24}} & \underline{\textbf{5.56}} & \underline{\textbf{189.2}} & \underline{\textbf{0.84}} & \underline{\textbf{0.38}} \\
+MG$_\textup{(ours)}$ & 0.15 & 7.62 & 5.99 & 183.4 & 0.83 & 0.38 \\
+MG$_\textup{(ours)}$ & 0.20 & 9.01 & 7.04 & 171.7 & 0.81 & 0.36 \\
\midrule
SiT-B/2 & 1.00 & 33.0 & 27.8 & 65.24 & 0.68 & 0.35 \\
+MG$_\textup{(ours)}$ & 0.05 & 10.8 & 9.25 & 168.8 & 0.80 & 0.34 \\
+MG$_\textup{(ours)}$ & 0.10 & \underline{\textbf{6.49}} & \underline{\textbf{5.69}} & \underline{\textbf{212.3}} & \underline{\textbf{0.86}} & \underline{\textbf{0.38}} \\
+MG$_\textup{(ours)}$ & 0.15 & 6.77 & 5.89 & 207.4 & 0.85 & 0.37 \\
+MG$_\textup{(ours)}$ & 0.20 & 8.87 & 8.06 & 199.6 & 0.84 & 0.37 \\
\bottomrule
\end{tabular}
}%
\end{sc}
\end{small}
\end{center}
\vskip -0.1in
\end{table}

\subsection{Overall Performances}
\label{sec:overall_performance}

First of all, we present a through system-level comparison with recent state-of-the-art image generation approaches on ImageNet $256\times256$ dataset in \cref{tab:imagenet-256-wo-cfg,tab:imagenet-256-w-cfg}. As shown in \cref{tab:imagenet-256-wo-cfg}, both DiT-XL/2 and SiT-XL/2 models greatly benefit from our method, achieving the outstanding performance gain of $78.9\%$ and $84.4\%$. It is worth mentioning that our models do not apply modern techniques in the inference process, including rejection sampling~\cite{tian2024visual}, classifier-free guidance~\cite{ho2020denoising} and guidance interval~\cite{kynkaanniemi2024applying}. Compared to advanced methods, our models are light-weight, \eg 675M in contrast to RAR-XXL with 1.5B and MAR-H with 943M parameters, and consume less computational resources, for example, LightningDiT uses DiT-XL/1 to reduce patch size to $1\times1$ and needs $16\times$ computation in attention operations.

To facilitate a fair evaluation, we also compare with other methods with Classifier-free guidance. While prevalent diffusion models significantly benefit and are indispensable from CFG, it introduces an additional forward without condition and doubles the computation consumptions. Also, it usually requires a careful search over the hyper-parameter of guidance scale to achieve the best trade-off between quality and diversity. In contrast, our models still surpass other CFG-assisted methods and run with only half of the generation time.

Finally, we report the time consumption for each model to generate one sample in seconds. Comparing to other diffusion-based approaches facilitated with vanilla CFG, our method runs significantly faster and does not sacrifice inference speed for sampling quality.

\begin{table}[t]
\caption{Experiments on Model input $w$.}
\label{tab:ablation-input-scale}
\begin{center}
\begin{small}
\begin{sc}
\vskip -0.2in
\scalebox{0.8}{
\begin{tabular}{lc|ccccc}
\toprule
Model & $w$-in & FID$\downarrow$   & sFID$\downarrow$  & IS$\uparrow$     & Pre.$\uparrow$ & Rec.$\uparrow$ \\
\midrule
DiT-B/2 & \ding{55} & 43.5 & 36.7 & 39.23 & 0.62 & 0.34 \\
+MG$_\textup{(ours)}$ & \ding{55} & \underline{\textbf{7.24}} & \underline{\textbf{5.56}} & \underline{\textbf{189.2}} & \underline{\textbf{0.84}} & 0.38 \\
+MG$_\textup{(ours)}$ & \ding{51} & 8.13 & 6.03 & 175.1 & 0.84 & \underline{\textbf{0.39}} \\
\midrule
SiT-B/2 & \ding{55} & 33.0 & 27.8 & 65.24 & 0.68 & 0.35 \\
+MG$_\textup{(ours)}$ & \ding{55} & \underline{\textbf{6.49}} & \underline{\textbf{5.69}} & \underline{\textbf{212.3}} & \underline{\textbf{0.86}} & \underline{\textbf{0.38}} \\
+MG$_\textup{(ours)}$ & \ding{51} & 7.33 & 5.96 & 207.4 & 0.85 & 0.38 \\
\bottomrule
\end{tabular}
}%
\end{sc}
\end{small}
\end{center}
\vskip -0.1in
\end{table}

\begin{table}[t]
\caption{Experiments on empty class $\varnothing$.}
\label{tab:ablation-empty-class}
\begin{center}
\begin{small}
\begin{sc}
\vskip -0.2in
\scalebox{0.8}{
\begin{tabular}{lc|ccccc}
\toprule
Model & $\varnothing$-cls & FID$\downarrow$   & sFID$\downarrow$  & IS$\uparrow$     & Pre.$\uparrow$ & Rec.$\uparrow$ \\
\midrule
DiT-B/2 & \ding{55} & 43.5 & 36.7 & 39.23 & 0.62 & 0.34 \\
+MG$_\textup{(ours)}$ & \ding{55} & 9.66 & 8.73 & 174.4 & 0.81 & 0.35 \\
+MG$_\textup{(ours)}$ & \ding{51} & \underline{\textbf{7.24}} & \underline{\textbf{5.56}} & \underline{\textbf{189.2}} & \underline{\textbf{0.84}} & \underline{\textbf{0.38}} \\
\midrule
SiT-B/2 & \ding{55} & 33.0 & 27.8 & 65.24 & 0.68 & 0.35 \\
+MG$_\textup{(ours)}$ & \ding{55} & 9.03 & 7.96 & 183.3 & 0.82 & 0.35 \\
+MG$_\textup{(ours)}$ & \ding{51} & \underline{\textbf{6.49}} & \underline{\textbf{5.69}} & \underline{\textbf{212.3}} & \underline{\textbf{0.86}} & \underline{\textbf{0.38}} \\
\bottomrule
\end{tabular}
}%
\end{sc}
\end{small}
\end{center}
\vskip -0.1in
\end{table}

\begin{table}[t]
\captionsetup{width=0.85\columnwidth}
\caption{Experiments on Model size. Our method scales to models with different sizes.}
\label{tab:model-size}
\begin{center}
\begin{small}
\begin{sc}
\vskip -0.2in
\scalebox{0.8}{
\begin{tabular}{l|ccccc}
\toprule
Model & FID$\downarrow$   & sFID$\downarrow$  & IS$\uparrow$     & Pre.$\uparrow$ & Rec.$\uparrow$ \\
\midrule
DiT-B/2  & 43.5 & 36.7 & 39.23 & 0.62 & 0.34 \\
+MG$_\textup{(ours)}$ & 7.24 & 5.56 & 189.2 & 0.84 & 0.38 \\
\midrule
DiT-L/2  & 23.3 & 18.4 & 132.7 & 0.73 & 0.40 \\
+MG$_\textup{(ours)}$ & 5.43 & 4.66 & 236.3 & 0.83 & 0.44 \\
\midrule
DiT-XL/2  & 19.5 & 15.6 & 163.5 & 0.79 & 0.46 \\
+MG$_\textup{(ours)}$ & 3.37 & 4.73 & 257.2 & 0.84 & 0.51 \\
\midrule
SiT-B/2  & 33.0 & 27.8 & 65.24 & 0.68 & 0.35 \\
+MG$_\textup{(ours)}$ & 6.49 & 5.69 & 212.3 & 0.86 & 0.38 \\
\midrule
SiT-L/2  & 18.9 & 16.3 & 173.2 & 0.71 & 0.42 \\
+MG$_\textup{(ours)}$ & 4.50 & 4.03 & 243.9 & 0.85 & 0.46 \\
\midrule
SiT-XL/2  & 17.3 & 13.9 & 192.1 & 0.78 & 0.50 \\
+MG$_\textup{(ours)}$ & 2.89 & 3.12 & 261.0 & 0.85 & 0.54 \\
\bottomrule
\end{tabular}
}%
\end{sc}
\end{small}
\end{center}
\vskip -0.1in
\end{table}

\begin{table}[t]
\captionsetup{width=0.85\columnwidth}
\caption{Experiments on ImageNet 512. Our method scales to high-resolution image datasets.}
\label{tab:imagenet-512}
\begin{center}
\begin{small}
\begin{sc}
\vskip -0.2in
\scalebox{0.8}{
\begin{tabular}{l|ccccc}
\toprule
Model & FID$\downarrow$   & sFID$\downarrow$  & IS$\uparrow$     & Pre.$\uparrow$ & Rec.$\uparrow$ \\
\midrule
DiT-XL/2 & 3.04 & 5.02 & 240.8 & 0.84 & 0.54 \\
+MG$_\textup{(ours)}$ & 2.78 & 4.86 & 257.2 & 0.83 & 0.58 \\
\midrule
SiT-XL/2 & 2.62 & 4.18 & 252.2 & 0.84 & 0.57 \\
+MG$_\textup{(ours)}$ & 2.24 & 4.03 & 276.9 & 0.86 & 0.60 \\
\bottomrule
\end{tabular}
}%
\end{sc}
\end{small}
\end{center}
\vskip -0.1in
\end{table}

\begin{figure}[t]
    \centering
    \begin{minipage}[!t]{0.5\linewidth}
        \centering
        \includegraphics[width=1\textwidth]{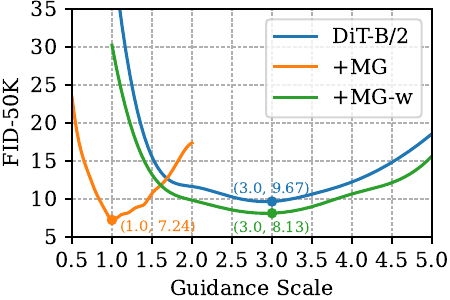}
    \end{minipage}%
    \begin{minipage}[!t]{0.5\linewidth}
        \centering
        \includegraphics[width=1\textwidth]{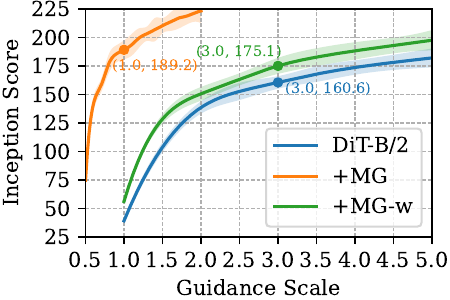}
    \end{minipage} \\
    \begin{minipage}[!t]{0.5\linewidth}
        \centering
        \includegraphics[width=1\textwidth]{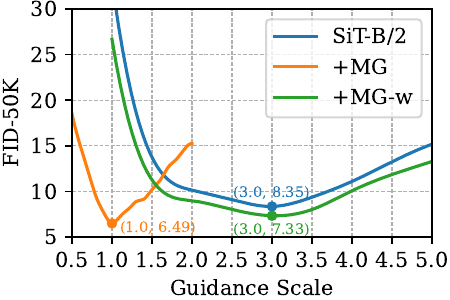}
    \end{minipage}%
    \begin{minipage}[!t]{0.5\linewidth}
        \centering
        \includegraphics[width=1\textwidth]{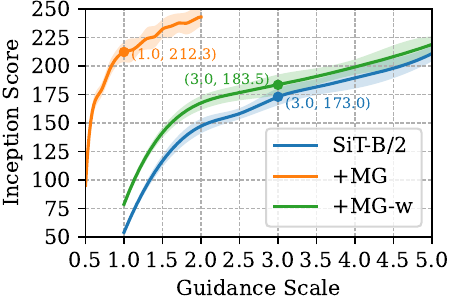}
    \end{minipage}
    \vskip -3mm
    \caption{FID-50K and Inception Score results as the guidance scale increases during inference. Our method is compatible with and can be wrapped into vanilla CFG.}
    \label{fig:fid-vs-cfg-scale}
\end{figure}

\begin{figure}[t]
    \centering
    \begin{minipage}[!t]{0.5\linewidth}
        \centering
        \includegraphics[width=1\textwidth]{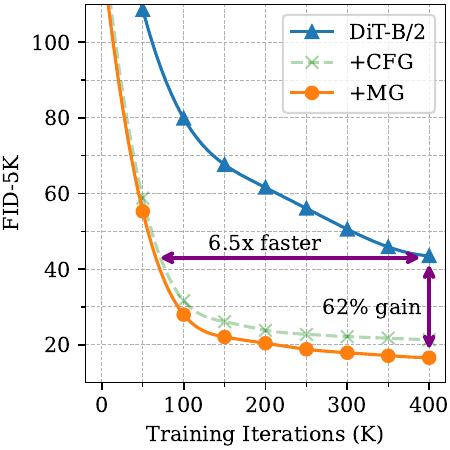}
    \end{minipage}%
    \begin{minipage}[!t]{0.5\linewidth}
        \centering
        \includegraphics[width=1\textwidth]{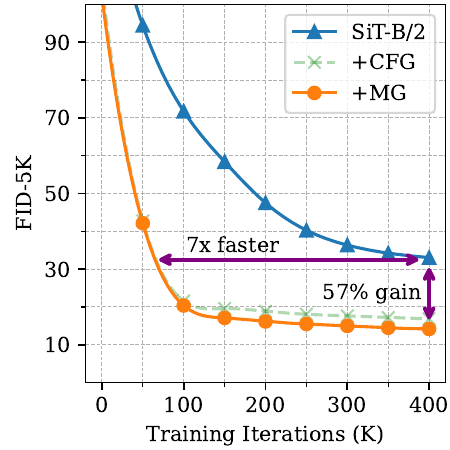}
    \end{minipage}
    \vskip -3mm
    \caption{FID-5K results during training. Our method is $\ge6.5\times$ faster and $\approx60\%$ better than vanilla DiT and SiT, even surpassing the results of CFG.}
    \label{fig:training-speedup}
\end{figure}

\begin{figure}[!ht]
    \centering
    \begin{minipage}[!t]{0.5\linewidth}
        \centering
        \includegraphics[width=0.98\textwidth]{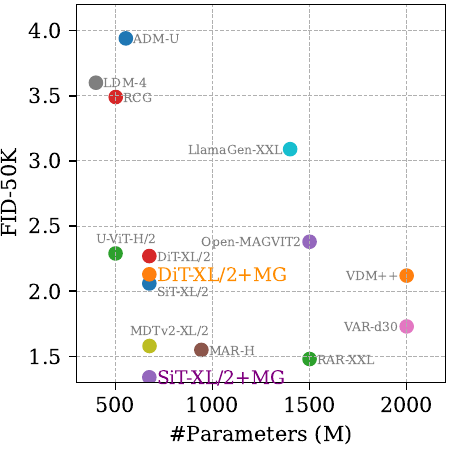}
    \end{minipage}%
    \begin{minipage}[!t]{0.5\linewidth}
        \centering
        \includegraphics[width=0.98\textwidth]{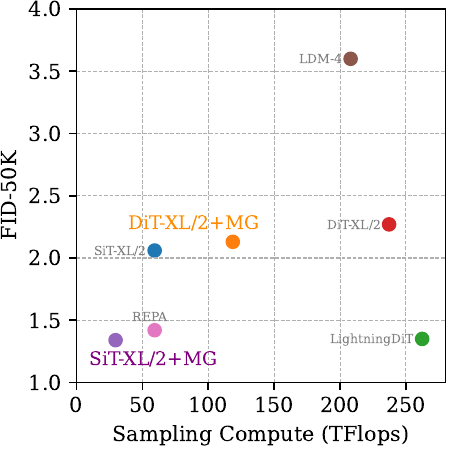}
    \end{minipage}
    \vskip -3mm
    \caption{FID-50K \vs\ number of parameters and sampling flops of different models, where our models are highlighted.}
    \label{fig:parameters-inference-speedup}
    \vskip -0.1in
\end{figure}

\subsection{Ablation study}
\label{sec:ablation_study}

To thoroughly understand the designs and subsequent influences of our method, we conduct ablation experiments on the key components, including the hyper-parameter $w$, $\lambda$ choices, whether the model takes $w$ as input, and the role of empty class during training. Moreover, we assess the scalability of our method in terms of both model size and dataset difficulty.

\noindent \textbf{Hyper-parameter $w$} In \cref{eq:mg_noise,eq:mg_flow}, the hyper-parameter $w$ controls the scale of posterior probability and serves an important role akin to the guidance scale in CFG, which is sensitive to FID-score. We conduct ablation experiments on the hyper-parameter $w$ and report results in \cref{tab:ablation-scale}, where $w=1$ refers to vanilla forms in DiT~\cite{peebles2023scalable} and SiT~\cite{ma2024sit}. It is shown that the choice of $w$ also acts as a crucial role and balances the trade-off between quality and diversity.

To overcome the tiresome and costly search of $w$ during training, we propose an adaptive approach to automatically adjust $w$, which achieves comparable performance with manual search. Meanwhile, we can further apply CFG to our models in \cref{fig:fid-vs-cfg-scale} to flexibly adjust between better quality and diversity during inference.

\noindent \textbf{Hyper-parameter $\lambda$} The relative ratio to train conditional and unconditional models, $\lambda$, is also important to our method. The unconditional model is usually trained by randomly dropping the condition and replacing with an additional empty label for part of training data. In \cref{tab:ablation-drop-ratio}, we conduct ablation experiments on the hyper-parameter $\lambda$ report the corresponding results. We find that $\lambda$ is less sensitive than $w$, and $\lambda\in\{0.10,0.15\}$ offers satisfactory performances.

\noindent \textbf{Model input $w$} Despite the same loss formulation in the \cref{eq:final_loss}, it is optional whether our model takes the scale $w$ as an additional input. In \cref{tab:ablation-input-scale}, the models with $w$-input slightly lag behind the counterparts without $w$-input but still exceeding the vanilla DiT-B/2 and SiT-B/2 with CFG, demonstrating the superiority of our method.

\noindent \textbf{Empty class $\varnothing$} In \cref{tab:ablation-empty-class}, we conduct ablation experiments on the introduction of additional empty class. While removing the empty class in our method leads to worse estimation of posterior probability, the generation performances are still on par with the vanilla CFG. It can also be improved by better estimation with the law of total probability or a larger batch size.

\begin{figure*}[t]
\begin{center}
\centerline{\includegraphics[width=1\linewidth]{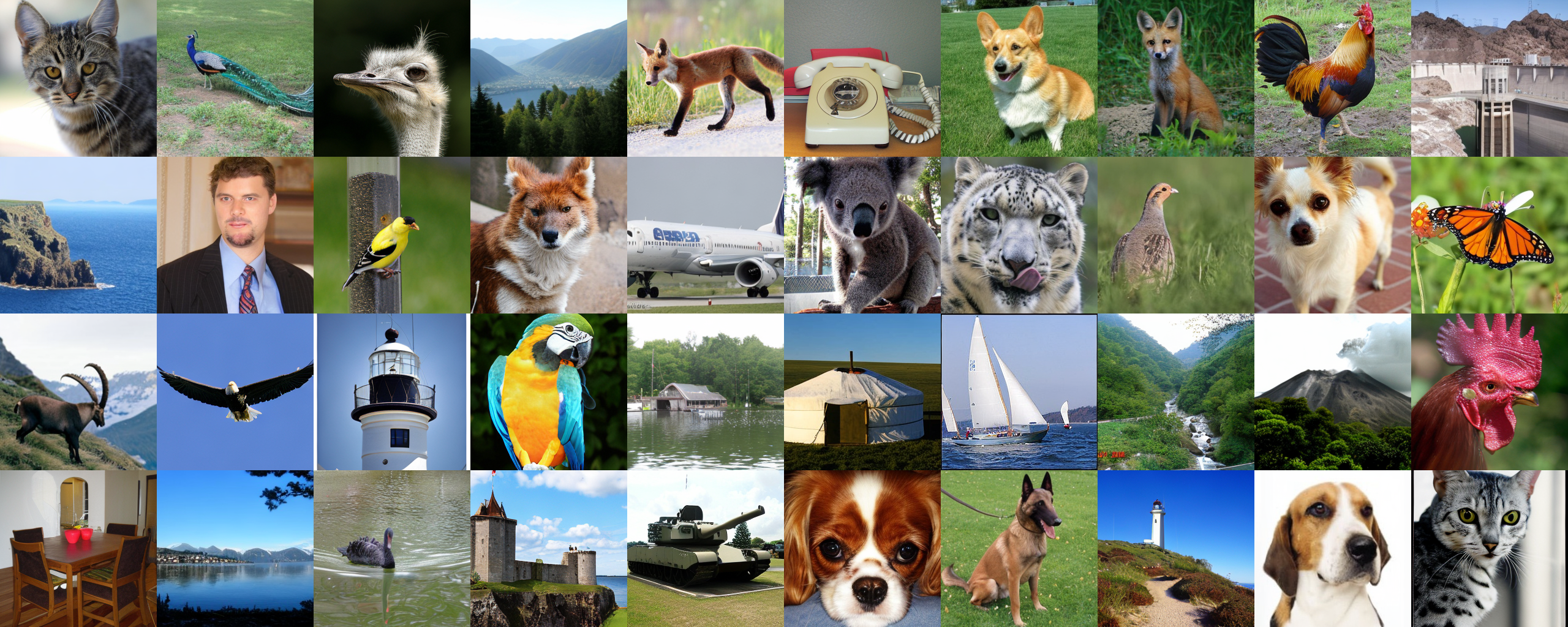}}
\caption{\textbf{Uncurated} samples of SiT-XL/2+MG on ImageNet $256\times256$.}
\label{fig:generated-images}
\end{center}
\vskip -0.2in
\end{figure*}

\begin{figure*}[!ht]
\begin{center}
\centerline{\includegraphics[width=1\linewidth]{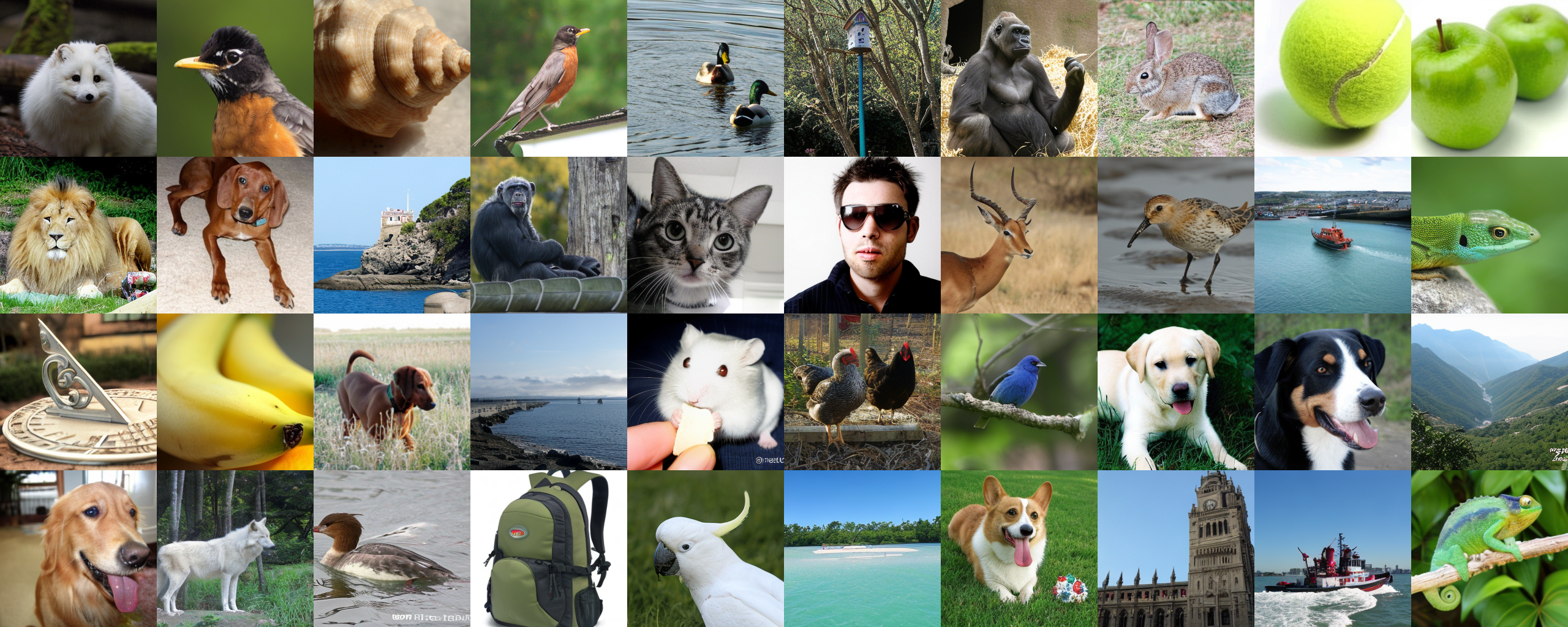}}
\caption{\textbf{Uncurated} samples of SiT-XL/2+MG on ImageNet $512\times512$.}
\label{fig:generated-images}
\end{center}
\vskip -0.2in
\end{figure*}

\noindent \textbf{Efficiency} One key advantage of our method is that it not only improves inference speed by avoiding the second network forward of CFG, but also accelerates the training and convergence of diffusion models. In \cref{fig:training-speedup}, our method obtains $\ge6.5\times$ convergence speed and $\approx60\%$ performance gain. In \cref{fig:parameters-inference-speedup}, we plot the number of network parameters and sampling compute in TFlops versus FID-50K of concurrent methods. When comparing number of network parameters, our method comes with the lowest FID and a small model size. When comparing sampling computes, our method achieves state-of-the-art performances in parallel with LightningDiT~\cite{yao2025reconstruction}, while requires only $\approx12\%$ computational resources.

\noindent \textbf{Scalability} Finally, the scalability to larger model and dataset of our method is of imparable significance. In \cref{tab:model-size}, we conduct ablation stuides on model size with B/2, L/2 and XL/2 variants of DiT and SiT models. It is demonstrated that our method is capable to boost the performance of models with different sizes and designs. We scale to ImageNet $512\times512$ dataset to validate our method in handling difficult distributions in~\cref{tab:imagenet-512}. As depicted, our method also offers improvements on high-resolution tasks.


\section{Conclusion}

This work addresses the limitations of the commonly used Classifier-free guidance (CFG) of diffusion models, and proposes Model-guidance (MG) as an efficient and advantageous replacement. We first investigate the mechanism of CFG and locate the source of performance gain as a joint optimization of posterior probability. Then, we transcend the idea into the training process of diffusion models and directly learn the score of the joint distribution, $\nabla\log \tilde{p}_\theta(x_t|c)=\nabla\log p_\theta(x_t|c)p_\theta(c|x_t)^w$. Comprehensive experiments demonstrate that our method significantly boosts the generation performance without efficiency loss, scales to different models and datasets, and achieves state-of-the-art results on ImageNet $256\times256$ dataset. We believe that this work contributes to future diffusion models.


\clearpage
\newpage

\section*{Impact Statements}

This paper propose methods in association with generative methods. There might be potential negative social impacts, \eg generating fake portraits, as the core contribution of our work is a new algorithm of generative modeling. As possible mitigation strategies, we will restrict the access to these models in the planned release of code and models. We also validate that current detectors can effectively determine our generation results about human portraits.


\bibliography{main_paper}
\bibliographystyle{icml2025}
\clearpage


\end{document}